\documentclass[runningheads]{llncs}

\usepackage[T1]{fontenc}
\usepackage[utf8]{inputenc}
\usepackage[english]{babel}

\usepackage{graphicx}
\usepackage{amsmath,amssymb}
\usepackage{booktabs}
\usepackage{tabularx}
\usepackage{hyperref}
\usepackage{color}
\usepackage{array}  

\usepackage{cite}
\usepackage{multirow}
\usepackage{adjustbox}

\begin{document}

\title{DIMT25@ICDAR2025: HW-TSC's End-to-End Document Image Machine Translation System Leveraging Large Vision-Language Model}
\titlerunning{DIMT25@ICDAR2025: HW-TSC's System Report}

\author{Zhanglin Wu \and Tengfei Song \and Ning Xie \and Weidong Zhang \and \\ Pengfei Li \and  Shuang Wu \and Chong Li \and Junhao Zhu  \and Hao Yang}
\authorrunning{Wu et al.}

\institute{
Huawei Translation Service Center, Nanjing, China \\ \email{\{wuzhanglin2, songtengfei2, nicolas.xie, zhangweidong17, \\lipengfei203, wushuang42, august.li, zhujunhao, yanghao30\}@huawei.com} 
}

\maketitle

\begin{abstract}

This paper presents the technical solution proposed by Huawei Translation Service Center (HW-TSC) for the "End-to-End Document Image Machine Translation for Complex Layouts" competition at the 19th International Conference on Document Analysis and Recognition (DIMT25@ICDAR2025). Leveraging state-of-the-art open-source large vision-language model (LVLM), we introduce a training framework that combines multi-task learning with perceptual chain-of-thought to develop a comprehensive end-to-end document translation system. During the inference phase, we apply minimum Bayesian decoding and post-processing strategies to further enhance the system's translation capabilities. Our solution uniquely addresses both OCR-based and OCR-free document image translation tasks within a unified framework. This paper systematically details the training methods, inference strategies, LVLM base models, training data, experimental setups, and results, demonstrating an effective approach to document image machine translation.
\end{abstract}

\section{Introduction}





Document Image Machine Translation (DIMT) \cite{ref_dimt1, ref_dimt2} combines text recognition and translation to process multilingual documents \cite{ref_dimt3} like manuals, reports, and archives. Despite deep learning \cite{ref_dl} advances, building an end-to-end DIMT system that handles diverse layouts with high accuracy remains challenging.

To foster innovation in this field, the 19th International Conference on Document Analysis and Recognition (ICDAR 2025) has specifically established a competition titled "End-to-End Document Image Machine Translation for Complex Layouts." Addressing this challenge, our team (HW-TSC) has developed an innovative framework based on cutting-edge large vision-language models (LVLMs) \cite{ref_lvlm1,ref_lvlm2}. This framework integrates multi-task learning (MTL) \cite{ref_mtl} with perceptual chain-of-thought \cite{ref_cot} (PCOT) training approach, enabling the model to simultaneously comprehend visual layouts and linguistic content. During the inference phase, the system further employs minimum Bayesian (MBR) decoding \cite{ref_mbr} and post-processing strategies to optimize output quality.

The primary innovation of this study lies in unifying OCR-based and OCR-free translation tasks within a single framework, eliminating the need for constructing independent pipelines. This design significantly enhances the system's adaptability, allowing it to flexibly handle real-world scenarios where Optical Character Recognition (OCR) \cite{ref_ocr} may or may not be applicable. Evaluations on the test set confirm the method's effectiveness in processing documents with complex structures.

This paper systematically presents the complete technical solution, emphasizing reproducibility and transparency. The subsequent sections elaborate in detail on: 1) the training methodology and inference strategies, 2) the composition of training data and the LVLM base model, 3) experimental setups and results. This structured presentation facilitates a comprehensive understanding of our research contributions and innovations.

\section{Method Description}



\subsection{Training Method}



\begin{figure}[htbp]
\begin{center}
\includegraphics[width=\textwidth, height=0.6\textwidth]{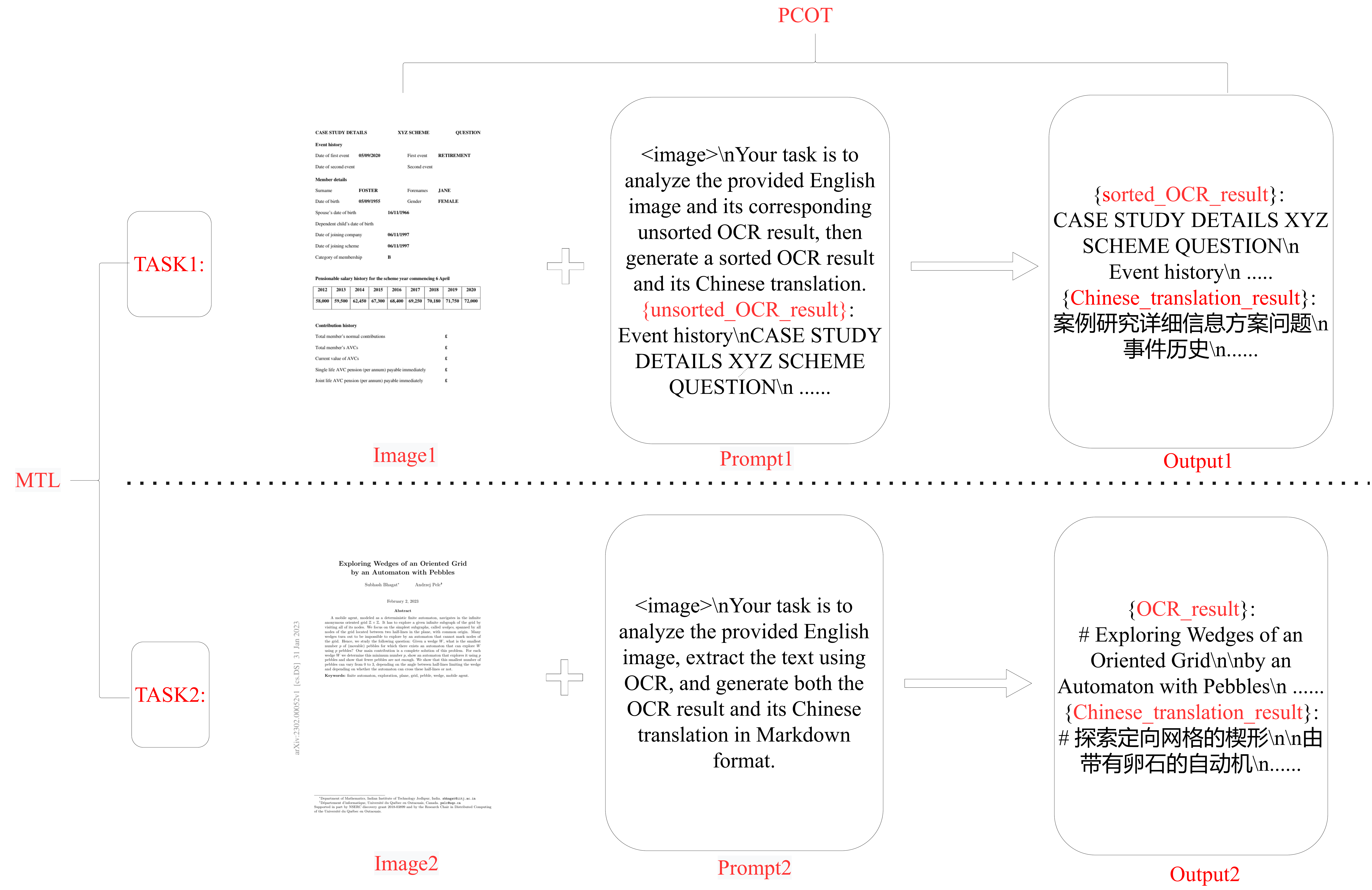}
\caption{Training Data Organization Structure for Our Method.}\label{fig:EX}
\end{center}
\vspace{-0.5cm}
\end{figure}

With the advancement of large language model (LLM) \cite{ref_llm1,ref_llm2} technology, its three-stage training paradigm of "Continue Pre-Training (CPT) \cite{ref_cpt}–Supervised Fine-Tuning (SFT) \cite{ref_sft}–Reinforcement Learning from Human Feedback (RLHF) \cite{ref_rlhf}" has been successfully applied to LVLMs \cite{ref_qwen,ref_gemini}. This paradigm enables open-source LVLMs to achieve outstanding performance in cross-modal tasks \cite{ref_cm}. Based on this, we propose an innovative training strategy: integrating MTL with PCOT to perform SFT on a state-of-the-art LVLM \cite{ref_lvlm2}, enhancing its performance on DIMT25 tasks. Figure \ref{fig:EX} illustrates the training data format.

\subsubsection{MTL}


MTL \cite{ref_mtl} is a machine learning paradigm  \cite{ref_mlp} that enhances model generalization by sharing representations across different tasks. Specifically, MTL concurrently optimizes multiple related tasks within a single model, capitalizing on latent inter-task correlations to effectively mitigate issues prevalent in traditional single-task learning \cite{ref_stl}—such as model overfitting or limited representational capacity caused by data isolation. Particularly during the SFT process of LVLMs, MTL facilitates cross-task collaborative training through parameter-sharing mechanisms and joint loss function optimization. This approach not only strengthens the model's ability to extract multimodal features but also markedly improves its robustness in semantic understanding within complex scenarios.

\subsubsection{PCOT}


Chain-of-Thought (CoT) \cite{ref_cot} approach demonstrates significant advantages in complex cognitive tasks (such as mathematical reasoning, logical deduction, and multimodal comprehension) by simulating the progressive reasoning process of humans. Compared to traditional end-to-end training paradigms, CoT emphasizes an explicit step-by-step reasoning mechanism, with this structured processing enhancing the transparency of model decision-making. To address the unique challenges of the DIMT25 task, we extend the standard CoT into PCOT. It adopts a hierarchical processing pipeline: the first stage focuses on precise detection and recognition of textual content in images, while the second stage performs specialized cross-lingual transformation. This two-stage design effectively resolves the semantic fragmentation issue of traditional methods in mixed text-image scenarios. By establishing a deep coupling mechanism between visual perception and language understanding, it significantly improves the accuracy and consistency of translation results. 

\subsection{Inference Strategy}


\subsubsection{MBR Decoding}


The core objective of MBR decoding \cite{ref_mbr} is to select the optimal output by optimizing an expected utility function, which quantifies the similarity between candidate hypotheses and reference texts. Previous research \cite{ref_mbr1,ref_mbr2,ref_mbr3} has demonstrated the effectiveness of MBR decoding in machine translation tasks, where automatic evaluation metrics \cite{ref_comet20,ref_comet22} are typically employed as similarity measures. We adapt this approach to the DIMT25 task. Specifically, we concurrently collect candidate outputs generated by two decoding strategies—including deterministic outputs from beam search and 10 diverse samples produced by temperature and nucleus sampling (with $t=0.7$ and $p=0.95$). Subsequently, we compute pairwise similarity scores within the candidate set using BLEU \cite{ref_bleu} and select the sample with the highest similarity score as the final system output.

\subsubsection{Post-processing}


The analysis of the model's output reveals that when processing special symbols (such as -, …, \_, *, etc.) that appear repeatedly in images or content-heavy tables, the model tends to over-translate. To address this, we establish two simple processing rules: reducing consecutive special symbols exceeding 10 instances to 10 and removing translation outputs for overly complex tables. Additionally, for Chinese translation results, we implement continuous space normalization after jieba tokenization, replacing multiple consecutive spaces with a single space to make the output more aligned with human reading habits.

\section{Implementation Details}

\subsection{Training Data}


The DIMT25 competition's multimodal dataset, shown in Table \ref{tab:statistics}, serves as our data source for model training and evaluation, containing two independent tracks for different task scenarios. 


\subsection{LVLM Base Model}

We select InternVL2.5-8B-MPO\footnote{\url{https://huggingface.co/OpenGVLab/InternVL2_5-8B-MPO}} (large model) and InternVL2.5-1B-MPO\footnote{\url{https://huggingface.co/OpenGVLab/InternVL2_5-1B-MPO}} (small model) as our LVLM base models in different model size segments, both of which demonstrated outstanding performance on open-source LVLM evaluation benchmarks like MOTBench \cite{ref_motbench}, and perform SFT on them.

\section{Experimental}

\subsection{Setup}


We employ the open-source InternVL framework\footnote{\url{https://github.com/OpenGVLab/InternVL}} to perform full-parameter fine-tuning on the InternVL2.5-MPO model, utilizing DeepSpeed's zero3-offload technique to accelerate the training process. The detailed training configuration is as follows: we conduct parallel training across 8 GPUs with a batch size of 4 per GPU and a gradient accumulation step of 4. The initial learning rate is set to 4e-5. The maximum sequence length is configured to 8,192 tokens, and the model is trained for 5 epochs.

\subsection{Result}



Table \ref{tab:results} presents three key findings: (1) Compared to single-task direct SFT, the MTL-PCOT combined SFT method demonstrates more significant performance improvement for LVLM on the DIMT25 task; (2) MBR and post-processing inference strategies offer complementary performance enhancements; (3) Model capability scales with parameter size, with larger models outperforming smaller counterparts when using identical methodologies.

\begin{table}[htbp]
\centering
\caption{BLEU scores for each fine-grained DIMT25 sub-track}
\label{tab:results}
\begin{adjustbox}{width=\textwidth,height=0.22\textwidth,center}
\begin{tabular}{lccccccc}
\toprule
\multirow{2}{*}{Model} & \multicolumn{4}{c}{track1} & \multicolumn{2}{c}{track2} \\
\cmidrule(lr){2-5} \cmidrule(lr){6-7}
 & Valid-OCR & Valid-MT & Test-OCR & Test-MT & Valid-MT & Test-MT \\
\midrule
InternVL2.5-1B-MPO Zero-Shot & / & / & / & / & / & / \\
InternVL2.5-1B-MPO SFT & / & 67.21 & / & / & 59.81 & / \\
InternVL2.5-1B-MPO MTL-PCOT SFT & / & 70.81 & 94.63 & 62.16 & 62.17 & 57.35 \\
\hspace{1em}+ MBR & / & / & 96.50 & 64.08 & / & 59.06 \\
\hspace{1em}+ Post-processing & / & / & 97.00 & 66.16 & / & 59.56 \\
\midrule
InternVL2.5-8B-MPO Zero-Shot & / & 30.71 & / & / & 29.53 & / \\
InternVL2.5-8B-MPO SFT & / & 72.74 & / & / & 64.24 & / \\
InternVL2.5-8B-MPO MTL-PCOT SFT & / & 75.72 & 94.89 & 65.32 & 65.77 & 58.57 \\
\hspace{1em}+ MBR & / & / & 97.16 & 68.26 & / & 60.33 \\
\hspace{1em}+ Post-processing & / & / & 97.66 & 70.48 & / & 60.78 \\
\bottomrule
\end{tabular}
\end{adjustbox}
\end{table}

\vspace{1em}
\noindent \textbf{Compliance Statement:} \\
Our team confirms that all data used complies with the competition rules and relevant data privacy, copyright, and ethical standards.

\clearpage

\appendix

\section{Training Data Format}

\section{Training Data Size}

\begin{table}[h]
\centering
\caption{Statistical Information of the DIMT25 Training Data}
\tiny
\begin{adjustbox}{width=\textwidth,height=0.15\textwidth,center}
\begin{tabular}{lllllll lllllll ccccc ccccc c}
\toprule
Track &&&&&&& Dataset&&&&&&& \multicolumn{11}{c}{\# of Examples} \\
\cmidrule(lr){14-25}
 &&&&&&&  &&&&&&& Train &&&&& Valid &&&&& Test \\
\midrule
Track 1 &&&&&&& DIMT-WebDoc-300K &&&&&&& 300K &&&&& 1K &&&&& 1K \\
Track 2 &&&&&&& DIMT-arXiv-124K &&&&&&& 124K &&&&& 1K &&&&& 1K \\
\bottomrule
\end{tabular}
\end{adjustbox}
\label{tab:statistics}
\end{table}


\begin{thebibliography}{8}
\bibitem{ref_dimt1}
Zhang Z, Zhang Y, Liang Y, et al. LayoutDIT: Layout-aware end-to-end document image translation with multi-step conductive decoder[C]//Findings of the Association for Computational Linguistics: EMNLP 2023. 2023: 10043-10053.

\bibitem{ref_dimt2}
Liang Y, Zhang Y, Ma C, et al. Document image machine translation with dynamic multi-pre-trained models assembling[C]//Proceedings of the 2024 Conference of the North American Chapter of the Association for Computational Linguistics: Human Language Technologies (Volume 1: Long Papers). 2024: 7084-7095.

\bibitem{ref_dimt3}
Thendral, R., et al. "Document Image Analysis for Text Extraction and Translation." 2023 4th International Conference on Intelligent Technologies (CONIT). IEEE, 2024.

\bibitem{ref_dl}
LeCun, Yann, Yoshua Bengio, and Geoffrey Hinton. "Deep learning." nature 521.7553 (2015): 436-444.

\bibitem{ref_lvlm1}
Chen, Zhe, et al. "Expanding performance boundaries of open-source multimodal models with model, data, and test-time scaling." arXiv preprint arXiv:2412.05271 (2024).

\bibitem{ref_lvlm2}
Wang, Weiyun, et al. "Enhancing the reasoning ability of multimodal large language models via mixed preference optimization." arXiv preprint arXiv:2411.10442 (2024).

\bibitem{ref_mtl}
Zhang, Yu, and Qiang Yang. "A survey on multi-task learning." IEEE transactions on knowledge and data engineering 34.12 (2021): 5586-5609.

\bibitem{ref_cot}
Wei, Jason, et al. "Chain-of-thought prompting elicits reasoning in large language models." Advances in neural information processing systems 35 (2022): 24824-24837.


\bibitem{ref_mbr}
Farinhas, António, José GC de Souza, and André FT Martins. "An empirical study of translation hypothesis ensembling with large language models." arXiv preprint arXiv:2310.11430 (2023).

\bibitem{ref_ocr}
Mithe, Ravina, Supriya Indalkar, and Nilam Divekar. "Optical character recognition." International journal of recent technology and engineering (IJRTE) 2.1 (2013): 72-75.

\bibitem{ref_llm1}
Chang, Yupeng, et al. "A survey on evaluation of large language models." ACM transactions on intelligent systems and technology 15.3 (2024): 1-45.

\bibitem{ref_llm2}
Kasneci, Enkelejda, et al. "ChatGPT for good? On opportunities and challenges of large language models for education." Learning and individual differences 103 (2023): 102274.

\bibitem{ref_cpt}
Gupta, Kshitij, et al. "Continual pre-training of large language models: How to (re) warm your model?." arXiv preprint arXiv:2308.04014 (2023).

\bibitem{ref_sft}
Dong, Guanting, et al. "How abilities in large language models are affected by supervised fine-tuning data composition." arXiv preprint arXiv:2310.05492 (2023).

\bibitem{ref_rlhf}
Bai, Yuntao, et al. "Training a helpful and harmless assistant with reinforcement learning from human feedback." arXiv preprint arXiv:2204.05862 (2022).

\bibitem{ref_qwen}
Bai, Shuai, et al. "Qwen2. 5-vl technical report." arXiv preprint arXiv:2502.13923 (2025).

\bibitem{ref_gemini}
Team, Gemini, et al. "Gemini: a family of highly capable multimodal models." arXiv preprint arXiv:2312.11805 (2023).

\bibitem{ref_cm}
Shin, Andrew, Masato Ishii, and Takuya Narihira. "Perspectives and prospects on transformer architecture for cross-modal tasks with language and vision." International journal of computer vision 130.2 (2022): 435-454.

\bibitem{ref_mlp}
Lampropoulos, Aristomenis S., and George A. Tsihrintzis. "Machine learning paradigms." Applications in recommender systems. Switzerland: Springer International Publishing (2015).

\bibitem{ref_stl}
Marquet, Thomas, and Elisabeth Oswald. "A comparison of multi-task learning and single-task learning approaches." International Conference on Applied Cryptography and Network Security. Cham: Springer Nature Switzerland, 2023.


\bibitem{ref_mbr1}
Wu, Zhanglin, et al. "HW-TSC’s Submission to the CCMT 2024 Machine Translation Tasks." China Conference on Machine Translation. Singapore: Springer Nature Singapore, 2024.

\bibitem{ref_mbr2}
Wu, Zhanglin, et al. "Choose the Final Translation from NMT and LLM hypotheses Using MBR Decoding: HW-TSC's Submission to the WMT24 General MT Shared Task." arXiv preprint arXiv:2409.14800 (2024).

\bibitem{ref_mbr3}
Wu, Zhanglin, et al. "Improving the Quality of IWLST 2024 Cascade Offline Speech Translation and Speech-to-Speech Translation via Translation Hypothesis Ensembling with NMT models and Large Language Models." Proceedings of the 21st International Conference on Spoken Language Translation (IWSLT 2024). 2024.

\bibitem{ref_comet20}
Rei, Ricardo, et al. "COMET: A Neural Framework for MT Evaluation." Proceedings of the 2020 Conference on Empirical Methods in Natural Language Processing (EMNLP). 2020.

\bibitem{ref_comet22}
Rei, Ricardo, et al. "COMET-22: Unbabel-IST 2022 submission for the metrics shared task." Proceedings of the Seventh Conference on Machine Translation (WMT). 2022.

\bibitem{ref_bleu}
Papineni, Kishore, et al. "Bleu: a method for automatic evaluation of machine translation." Proceedings of the 40th annual meeting of the Association for Computational Linguistics. 2002.

\bibitem{ref_motbench}
Wu, Zhanglin, et al. "Evaluating Menu OCR and Translation: A Benchmark for Aligning Human and Automated Evaluations in Large Vision-Language Models" arXiv preprint arXiv:2504.13945 (2025).


\end{thebibliography}
\end{document}